\documentclass[runningheads]{llncs}
\usepackage[T1]{fontenc}
\usepackage{booktabs}
\usepackage{multirow}

\usepackage{graphicx}
\usepackage{amsmath}
\usepackage{float}
\usepackage{url}
\usepackage{hyperref}

\begin{document}
\title{DINOv2 for PET: A Pre-Training Domain Benchmark Across Classification, Segmentation, and Prognosis}

\titlerunning{PET-DINOv2-2D}

\author{Roxanne De Maeyer\inst{1,2,3}\and
Yannick Brackenier\inst{1} \and
Vicky De Ridder\inst{1} \and
Simon DeKeyser\inst{1} \and
Maarten Larmuseau\inst{1} \and
Jef Vandemeulebroucke\inst{2,3}}


%
\authorrunning{R. De Maeyer et al.}
%
\institute{Nuclivision bv, Edward Pynaertkaai 87, 9000 Gent, Belgium  \and
Vrije Universiteit Brussel (VUB), Pleinlaan 2, 1050 Brussel, Belgium   
\and
imec, Kapeldreef 75, 3001 Leuven, Belgium}
\maketitle       

\begin{abstract}

DINO (self-DIstillation with NO labels) has become a standard for self-supervised pre-training, producing features that generalise across datasets and downstream tasks. In medical imaging it has been adopted mainly for radiology (X-ray, CT and MRI), where even natural-image DINO features transfer well. 

In the literature, these features have been reported to transfer worse to functional imaging such as PET. Yet PET-specific DINO models have barely been introduced, and even less validated across multiple tasks and datasets.

We present PET-DINOv2-2D, the first DINOv2 model pre-trained on 2D PET, and benchmark it against open-source encoders: the original DINOv2 (pre-trained on the LVD-142M natural-image dataset) and OmniRad (pre-trained on CT, MRI and ultrasound). Using the same lightweight heads for every backbone, and despite far less pre-training data, PET-DINOv2-2D matches or outperforms both on three downstream PET tasks: normal/abnormal classification, lesion segmentation and recurrence-free survival prognosis. Its advantage is largest when few labels are available, suggesting that PET pre-training gives more transferable, label-efficient features. Code is made available at \url{https://github.com/nuclivision/pet-dinov2-2d/}.

\end{abstract}
\section{Introduction}
Self-supervised learning (SSL) has produced the first medical-imaging foundation models: a frozen encoder serving many downstream tasks through a small head. One pre-training objective usually suits only one task family: Wald et al.~\cite{waldOpenMind3DMedical2025} show contrastive pre-training favours classification while generative pre-training favours segmentation. DINO~\cite{caronEmergingPropertiesSelfSupervised2021}, and especially DINOv2~\cite{oquabDINOv2LearningRobust2024}, addresses this by learning features for both global and dense tasks.
Baharoon et al.~\cite{baharoonEvaluatingGeneralPurpose2024} confirm this: a frozen natural-image DINOv2 matches or beats every baseline on X-ray, CT and MRI, and is the only model to generalise across tasks.

PET is primarily an oncological modality: its functional uptake signal is used to detect, delineate and assess tumours, making a general-purpose PET foundation model particularly valuable. However, this cross-modality transfer breaks down on PET: natural-image DINO features that excel on X-ray/CT/MRI lose ground on PET/CT, attributed to PET's functional rather than structural signal~\cite{liuDoesDINOv3Set2026}. Pre-training DINO directly on PET has barely been explored: to our knowledge only two works do so~\cite{linNaturalImagesDualStreama,khatriExplainableVisionTransformer2023}, each on a single task, and Lin et al.\ train on UDPET~\cite{kuangyuUDPETUltralowDose2025} alone ($\approx1500$ scans, 2 scanners). Other SSL PET encoders likewise report gains on only one task~\cite{huangC2MAOTCrossmodalComplementary,pengAutomaticTumorSegmentation2022}.

\begin{figure}[h!]
  \centering
  \includegraphics[width=\linewidth]{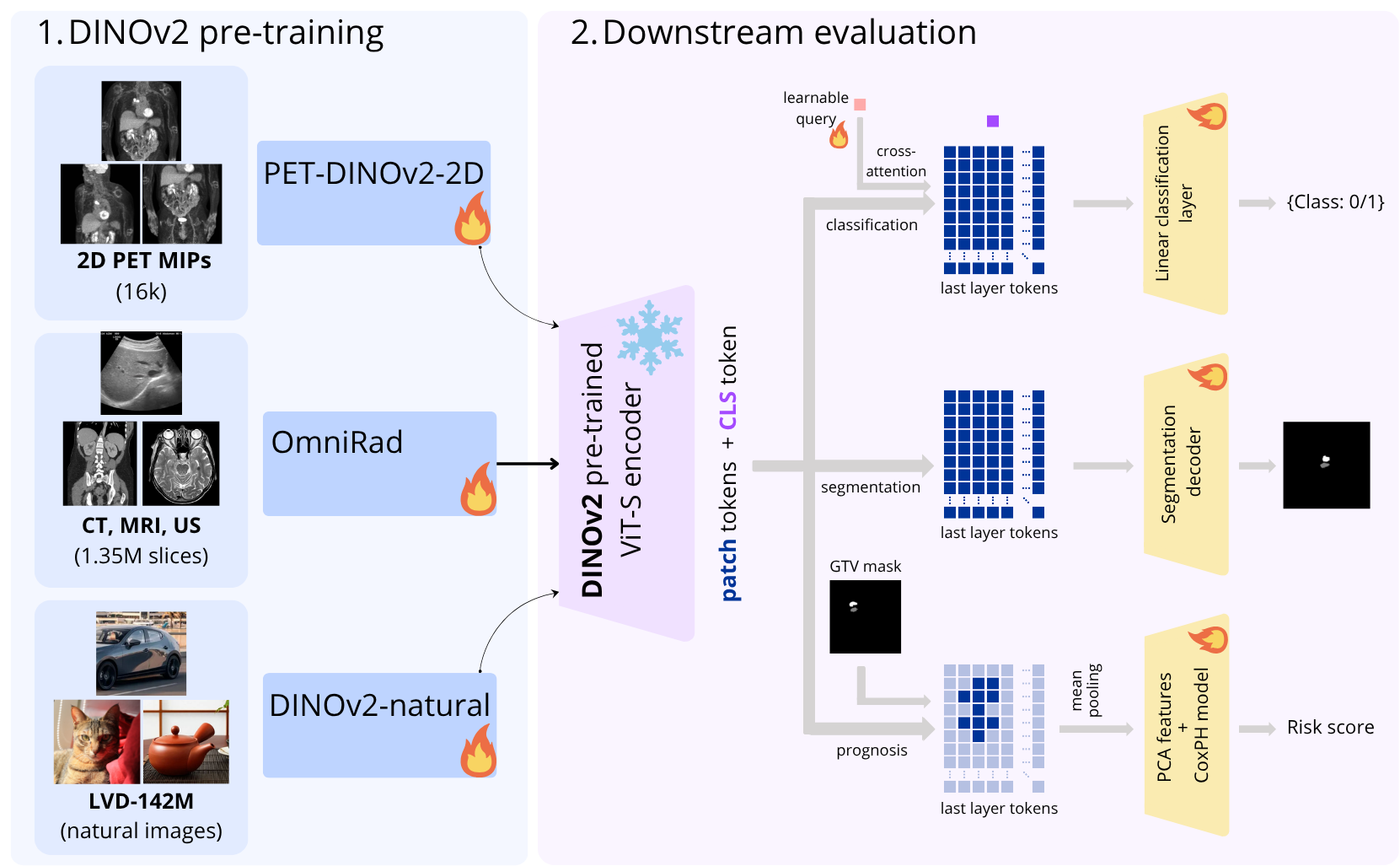}
  \caption{\textbf{Study overview.} We ask whether the pre-training \emph{domain} of a
  DINO encoder governs how well its frozen features transfer to PET. Three ViT-S encoders share the same architecture but differ in pre-training data: natural images (DINOv2), CT/MRI (OmniRad) and PET (our PET-DINOv2-2D). Each is used as a \emph{frozen}
  feature extractor for three downstream PET tasks: patient-level Normal/Abnormal
  classification, lesion segmentation, and recurrence-free survival prognosis. Every encoder
  is paired with the same lightweight task head.}
  \label{fig:downstream_summary}
\end{figure}

Testing whether this gap reflects a structural-vs-functional or merely a natural-vs-medical shift requires more than a single natural-image baseline. We therefore (1) pre-train our own DINOv2 on a dedicated, largely open-source PET dataset, and (2) benchmark natural-image, radiology (CT/MRI), and PET encoders across three core oncological tasks: lesion triage (detection), tumour-burden segmentation (delineation), and recurrence-free survival (prognosis) (Fig.~\ref{fig:downstream_summary}).

\subsubsection{Our Contributions:}

\begin{itemize}
\item We present the first PET-pre-trained DINOv2 that is evaluated beyond one downstream task (on patient-level classification, dense prediction, and survival prognosis) and benchmarked head-to-head against natural-image and radiology (CT/MRI) foundation models under a single frozen-encoder protocol.
\item We pre-train on multiple PET datasets rather than a single source, a prerequisite for a model intended to generalise as a foundation model.
\item We show that strong downstream performance is achievable with a \emph{frozen} encoder and a lightweight head, on both global and dense tasks.
\end{itemize}

\subsection{DINOv2 pre-training and baselines}
DINO~\cite{caronEmergingPropertiesSelfSupervised2021} trains a student ViT against an
EMA teacher on the \texttt{CLS} tokens only, giving strong image-level but weaker dense
features. One of the additions of DINOv2~\cite{oquabDINOv2LearningRobust2024} is the iBOT~\cite{zhouIBOTImageBERT2022}
loss on masked patch tokens, helping the features serve both global and dense tasks. We build on DINOv2, as the newer DINOv3~\cite{simeoniDINOv32025} is designed for a billion-image, billion-parameter regime less suited to the smaller datasets common in medical imaging.

We benchmark PET-DINOv2-2D (our PET pre-trained DINOv2) against two frozen DINOv2 baselines, all with a ViT-S architecture: natural-image DINOv2~\cite{oquabDINOv2LearningRobust2024} trained on LVD-142M, further referred to as ``DINOv2-natural'', and
OmniRad~\cite{zeddaOmniRadRadiologicalFoundation2026}, a DINOv2 pre-trained on
RadImageNet~\cite{meiRadImageNetOpenRadiologic2022} ($1.35$M CT, MRI and ultrasound
slices). OmniRad is the CT/MRI counterpart that
lets us separate the radiology-vs-PET axis from the natural-vs-medical one. For all
tasks we deliberately keep encoders \emph{frozen} and heads lightweight, so any
difference reflects the pre-training data rather than model capacity. An important caveat is that the pre-training dataset size of OmniRad is only $\approx\frac{1}{100}$ of LVD-142M.

We match each encoder's input to its pre-training: PET-DINOv2-2D takes the single-channel slice at patch size $16$ as it is pre-trained on raw, normalised single-uptake value (SUV), whereas the patch-$14$ OmniRad and DINOv2-natural take a $\log$-SUV slice replicated to three channels, adjusting the PET long-tail intensity range similarly to that of natural or CT \& MRI images. In every case the slice is cropped to the nearest multiple of the encoder's patch size.

\subsection{PET pre-training data}
No large-scale open-source PET dataset comparable to CheXpert~\cite{irvinCheXpertLargeChest2019} or RadImageNet in radiology exists yet. We therefore train PET-DINOv2-2D on existing open-source PET datasets plus collected private data, all in accordance with the Declaration of Helsinki.
The datasets used for pre-training are listed in Tab. \ref{tab:pretrainingdata}.

\begin{table}[h!]
    \centering
    \begin{tabular}{lccc}
        \toprule
        \textbf{Dataset name} & \textbf{PET tracer} & \textbf{\# Subjects} & \textbf{\# Scans} \\
        \midrule
        ACRIN-NSCLC \cite{ACRINNSCLCFDGPET}                & FDG             & 242  & 435  \\
        AutoPET \cite{gatidis2022autopetfdgpetct} & FDG      & 1014  & 1014  \\
        HECKTOR \cite{hecktor2025} & FDG & 680 & 680 \\
        NSCLC-Radiogenomics \cite{NSCLCRADIOGENOMICS}      & FDG             & 211  & 240  \\
        UDPET \cite{kuangyuUDPETUltralowDose2025}          & FDG             & 1447 & 1447 \\
        \textit{Private datasets}                          & FDG, PSMA, DOTA & 378  & 378  \\
        \bottomrule
    \end{tabular}
    \caption{Pre-training data}
    \label{tab:pretrainingdata}
\end{table}
The complete pre-training data consists of 4194 3D PET volumes. To pre-train a 2D
model on them we use Maximum Intensity Projections (MIPs): 4 oblique MIPs per
volume, giving 16,776 2D images. To our knowledge, MIPs have not been used for SSL
pre-training before. We choose them for PET specifically: tumour uptake typically
stands out above healthy-organ background in a projection, MIPs keep one image per
patient so patient-level labels stay intact, and nuclear medicine physicians
routinely read MIPs for quick triage. A projection also concentrates signal: most of a single PET slice is low-uptake
background, whereas each MIP captures the whole volume's uptake. The main trade-off is a hiding effect, where a small lesion behind a high-uptake
organ can be projected away. Pre-training on individual axial, coronal and sagittal
slices avoids this, but has to handle the per-slice sparsity, which we leave as
future work.

Across pre-training and all three downstream tasks, PET volumes are resampled to
$2\,$mm ($3\,$mm for segmentation) isotropic resolution and cropped or padded to
$256\times256$ in-plane before 2D extraction (MIPs for classification and prognosis,
slices for segmentation). For classification and segmentation we additionally probe
data efficiency by training on nested subsets of each fold's training pool and report
performance against the labelled-data fraction.

\subsection{Classification}
\subsubsection{Data}
The downstream cohort is a private multicentre FDG-PET dataset with 1,911 patients from three clinical centres. All data was collected in accordance with the Declaration of Helsinki. This cohort is disjoint from the pre-training data (no shared patients or scans), so the classification comparison is free of pre-training leakage.

Each scan has a patient-level binary label (normal vs.\ abnormal); Table~\ref{tab:triagedataset} gives the distribution and image specifications per centre. Labels were assigned per centre, with negatives followed up for six months to rule out missed findings. We hold out a centre- and label-stratified test set of 191 patients (10\%), tune with 5-fold cross-validation on the rest, then re-train with 8 seeds on the full training set and report on the test set.
 
\begin{table}
\caption{Downstream classification dataset}\label{tab:triagedataset}
\begin{tabular}{l||l|l|l}
 
&  \textbf{Centre 1} & \textbf{Centre 2} & \textbf{Centre 3}\\ \hline
Dataset size (patients) &  521 & 692 & 698 \\
Number of abnormal patients (AN) & 238 (45.7\%) & 347 (50.1\%) & 349 (50.0\%) \\

\end{tabular}
\end{table}

\subsubsection{Architecture of the head}
On a frozen encoder (features cached per patient), the head consumes the full token
sequence per view (\texttt{CLS} + patch tokens, $(N{+}1)\times D$, $D{=}384$ for
ViT-S). Since N/AN triage can hinge on a small lesion that a global \texttt{CLS} averages
away, all tokens are kept and re-weighted:
\begin{itemize}
  \item \textbf{Per-view pooling:} cross-attention with one learnable latent query
    over the $N{+}1$ tokens ($8$ heads, attention dropout $0.3$), then a residual
    LayerNorm + MLP, giving one $D$-vector per view.
  \item \textbf{Patient pooling:} mean of the per-view descriptors over the MIP views.
  \item \textbf{Classifier:} $\text{Dropout}(0.5)$ + linear $D\!\to\!2$.
  \item \textbf{Training:} only the latent pooler and linear classifier are trained, using cross-entropy
    loss, AdamW~\cite{loshchilovDecoupledWeightDecay2019} at learning rate $5\times10^{-4}$ (constant, no warm-up), for $150$
    epochs at an effective batch of $32$. Head and optimisation are identical across
    all backbones, so performance differences reflect feature quality.
\end{itemize}

\subsection{Segmentation}
\subsubsection{Data}
We evaluate dense prediction on the DeepPSMA dataset~\cite{jacksonRepositoryAnnotatedPSMA2025},
segmenting total tumour burden on the PSMA-PET scans. From the $100$ labelled
patients we hold out a (randomly sampled, splitting used a fixed seed of 42) test set of $10$ and split the remaining $90$ into a
5-fold cross-validation. For each patient we build a \emph{labelled} pool (slices with $\geq1$ lesion voxel)
and an \emph{empty} pool (no lesion, mean signal $>0.2\times$ the brightest slice).
Training samples slices at random with $10\%$ lesion-free; testing steps through
every slice.

\subsubsection{Architecture of the head}
We chose Primus as a lightweight segmentation head: 
\textbf{Primus}~\cite{waldPrimusEnforcingAttention2026} is skip-free, decoding only from the deepest frozen
token grid ($\sim0.2$M parameters). All inputs are SUV-clipped to $[0,20]$. The head training is optimised with AdamW at a base learning rate of $1\times10^{-3}$ for an effective batch of $8$, a $10$-epoch linear warm-up followed by cosine annealing. The loss is an equal-weight combination of a dynamic cross-entropy and a focal-Tversky term ($\alpha{=}0.3,\beta{=}0.7,\gamma{=}0.75$).
Between-backbone differences are assessed by a paired Wilcoxon
signed-rank test over the $6{\times}5{=}30$ (data-fraction, fold) Dice pairs.

\subsection{Prognosis}
\subsubsection{Data}
We evaluate prognosis on the HECKTOR 2025 recurrence-free survival (RFS) task
~\cite{hecktor2025,andrearczyk2023hecktor}, predicting time to relapse from FDG-PET
MIPs. The usable set has $651$ patients, of whom $132$ relapsed and $519$ ($79.7\%$)
are censored. Given the limited data and the ranking nature of the task, we do not repeat the low-data sweep for prognosis (too few events per fold).

Each patient uses the same oblique sagittal MIPs as the other tasks,
encoded once by the frozen backbone. We fuse routine clinical covariates (age, sex,
treatment, HPV status, WHO performance status, tobacco and alcohol use) and the log
metabolic tumour volume from the GTV segmentation. Missing values (flagged with a binary missingness indicator) are median-imputed
from training-fold statistics.
\subsubsection{Architecture of the head}
Every backbone is evaluated with the same protocol on its frozen, cached token
features:
\begin{itemize}
  \item \textbf{Pooling:} GTV-masked mean over patch tokens, then mean over the MIP
    views, giving one vector per patient.
  \item \textbf{Reduction + fusion:} PCA ($16$ components) on the image features, with or without
    concatenated z-score-standardised clinical covariates and log-MTV.
  \item \textbf{Risk model:} a penalised Cox proportional-hazards model
    \cite{coxRegressionModelsLifeTables1972} trained on the neg. Cox partial
    log-likelihood, mapping the fused vector to a scalar log-risk.
  \item \textbf{Train/val split:} the PCA, clinical-feature standardisation, and
    missing-value imputation are all fit on the training fold only, then applied
    unchanged to the validation fold.
\end{itemize}
We use centre-grouped 5-fold cross-validation (no centre in both train and
validation), matching HECKTOR's unseen-centre test split. As censoring is heavy and
centre-dependent, we report Uno's IPCW C-index~\cite{unoCstatisticsEvaluatingOverall2011}
as primary metric, alongside Harrell's C~\cite{harrellEvaluatingYieldMedical1982}.
Each backbone yields one pooled out-of-fold risk score per patient, and
between-encoder differences are tested with a paired patient-level bootstrap
($B{=}2000$; $95\%$ CIs, two-sided $p$)~\cite{efronIntroductionBootstrap1993}.

\section{Results \& Discussion}
\subsection{Classification}
\begin{figure}[h!]
  \centering
  \includegraphics[width=\linewidth]{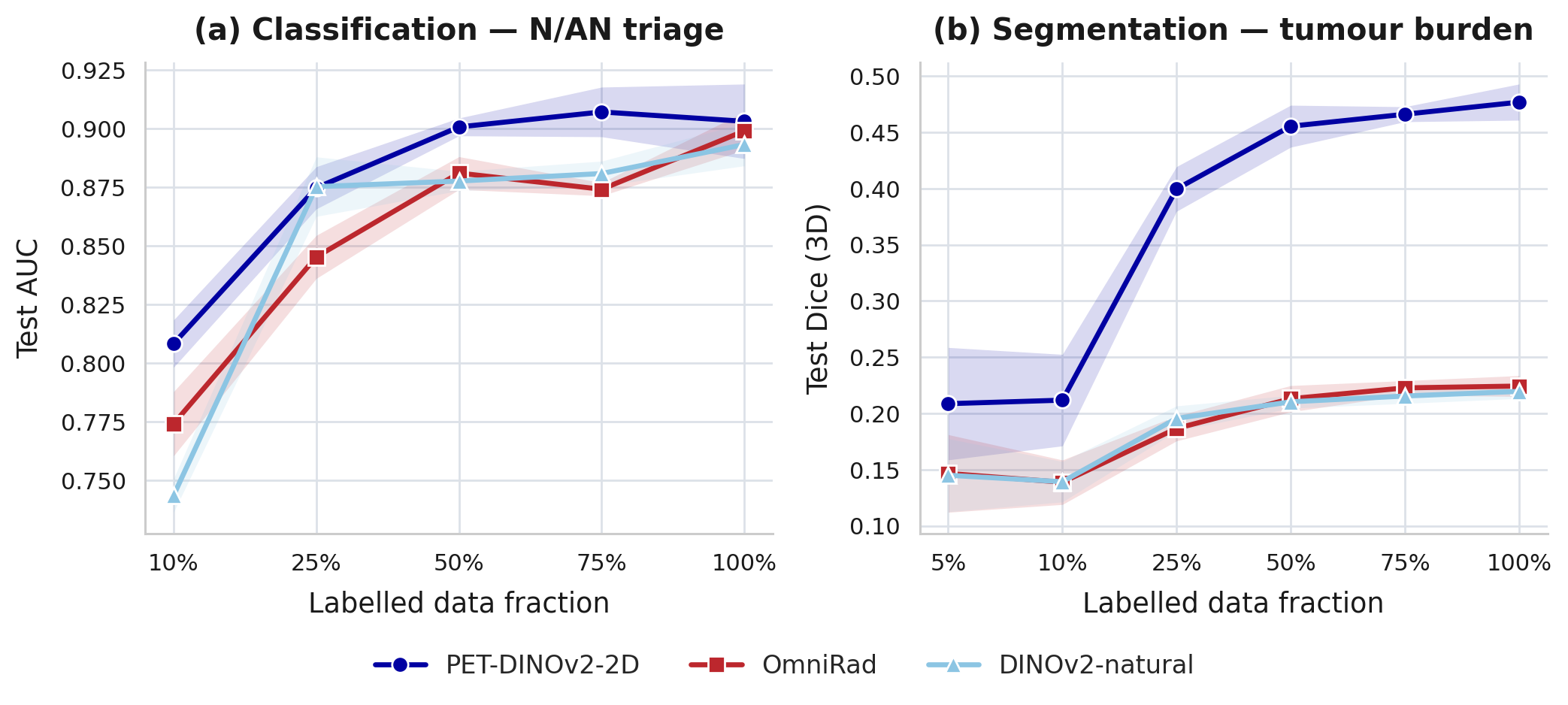}
  \caption{(a) Classification AUC vs. downstream data fraction on the triage PET-MIP task, (b) Held-out test Dice vs.\ labelled-data fraction on the DeepPSMA
  lesion task. Colour = backbone. Bands are $\pm1$\,SD over 8 training seeds in (a)
  and $\pm$std across the 5 cross-validation models in (b).}
  \label{fig:auc-seg-fraction}
\end{figure}

Figure~\ref{fig:auc-seg-fraction}(a) reports test AUC on the held-out triage set as a function of
the labelled-data fraction, each curve averaged over 8 training seeds (shaded bands:
$\pm1$\,SD). PET-DINOv2-2D attains the highest AUC at every fraction except 25\% and the full-data regime (100\%),
where it ties DINOv2-natural, and its margin is largest in the low-label regime
(\,$+0.055$ AUC at 10\%). Taking DINOv2-natural, which is the strongest, most widely used
baseline, as the pre-specified primary comparison, PET-DINOv2-2D shows a statistical significant improvement at 10\%, 50\%, and 75\% labelled data (paired Wilcoxon and $t$-test,
Holm-corrected, $p<0.05$). It likewise exceeds OmniRad at 25\%, 50\%, and 75\%
(paired $t$-test, Holm-corrected, $p\le0.01$; the Wilcoxon test is limited here by
the discreteness of $n{=}8$). All encoders converge at 100\% labelled data, where the differences are not significant. This is a first, soft confirmation our hypothesis: a PET-specific DINOv2, despite far less pre-training data, is a stronger feature extractor for PET tasks.

\subsection{Segmentation}

Figure~\ref{fig:auc-seg-fraction}(b) reports volumetric (3D, sagittal-stack) Dice on
DeepPSMA vs.\ labelled-data fraction. The skip-free Primus decodes only from the
deepest frozen tokens, so it depends \emph{entirely} on the features.

Under Primus, PET-DINOv2-2D dominates at every fraction, winning all $30/30$ paired
(fraction, fold) comparisons against both DINOv2-natural and OmniRad (Wilcoxon
signed-rank, $p<10^{-8}$): already at $5\%$ of the data the PET features carry
segmentation-ready structure the others lack. 

This contrast suggests that \emph{PET-ready} features are segmentation-useful under a
minimal head, where PET pre-training wins decisively. With this very lightweight head and frozen encoder features, Dice score results are generally low. Using more complex decoders could improve Dice scores.

\subsection{Prognosis}
The HECKTOR 2025 challenge \cite{hecktor2025} reported best held-out-centre RFS C-indices of $0.658$ and $0.602$ (first and second place), both from fully end-to-end 3D PET/CT encoders fused with clinical variables; imaging alone was deemed insufficient. In contrast, even \emph{frozen} 2D PET DINO features carry transferable prognostic signal under matched cross-centre evaluation. On centre-grouped folds (Table~\ref{tab:rfs}), image-only features from PET-DINOv2-2D are significantly more discriminative than those of DINOv2-natural ($\Delta$C$_{\mathrm{Uno}}=0.059$, $p=0.022$) and the radiology-pre-trained OmniRad ($\Delta$C$_{\mathrm{Uno}}=0.054$, $p=0.043$; both also significant under Harrell's C, $p=0.016$ and $0.028$). PET-DINOv2-2D is moreover the only encoder whose image-only discrimination is comparable to the clinical baseline (Uno C $0.617$ vs.\ $0.623$, overlapping 95\% CIs) and the only one whose fused image+clinical model exceeds clinical alone ($0.634$ vs.\ $0.623$), whereas natural-image and OmniRad features remain below the clinical reference.

\begin{table}[H]
\centering
\caption{Recurrence-free survival discrimination on HECKTOR (centre-grouped 5-fold CV;
PCA${=}16$). Image-only C-indices are pooled out-of-fold; 95\% CIs and $p$
(vs.\ PET-DINOv2-2D) are from a paired patient bootstrap. Best image-only value in
\textbf{bold}; $^{\ast}p<0.05$. Harrell-based $\Delta$\,($p$) vs.\ PET-DINOv2-2D: DINOv2-natural
$+.063$ (.016), OmniRad $+.058$ (.028).}
\label{tab:rfs}
\small
\setlength{\tabcolsep}{4.5pt}
\begin{tabular}{@{}lccc||c@{}}
\toprule
 & \multicolumn{2}{c}{Image only} & \begin{tabular}[t]{@{}c@{}}$\Delta$ vs.\ PET-DINOv2\\(image only)\end{tabular} & Img+clin.\\
\cmidrule(lr){2-3}
Encoder & Har. & Uno\,[95\% CI] & $\Delta$Uno ($p$) & Har. / Uno\\
\midrule
PET-DINOv2-2D & \textbf{0.610} & \textbf{0.617}\,[.566,.667] & --- & 0.622 / 0.634\\
OmniRad        & 0.552 & 0.564\,[.511,.618] & $-.054$ (\textbf{.043}$^{\ast}$) & 0.596 / 0.606\\
DINOv2-natural & 0.547 & 0.558\,[.505,.611] & $-.059$ (\textbf{.022}$^{\ast}$) & 0.584 / 0.596\\
\midrule
\multicolumn{5}{@{}l}{\textit{Clinical only (reference): Harrell 0.618, Uno 0.623 }}\\
\bottomrule
\end{tabular}
\end{table}

\section{Conclusion}
We introduced PET-DINOv2-2D, the first DINOv2 self-supervised on 2D PET, pre-trained on maximum-intensity projections from several PET datasets. Keeping every encoder frozen behind the same lightweight heads, we compared PET pre-training against the natural-image DINOv2 and radiology-pre-trained OmniRad on classification, segmentation and recurrence-free survival prognosis. PET pre-training gave the strongest features whenever labels were limited: it led classification in the low-data regime, outperformed both non-PET encoders on segmentation under a lightweight skip-free decoder, and gave the only image-only prognostic features matching the clinical baseline. OmniRad (CT/MRI) tracked natural-image DINOv2 throughout, while its pre-training dataset is also substantially smaller (factor 100). The transfer gap reflects overall a structural-vs-functional shift rather than a generic natural-vs-medical one.

Our study has three main limitations. First, HECKTOR (the prognosis set) was part of PET-DINOv2-2D's label-free pre-training but not the baselines', so prognosis is not leakage-free; we treat it as corroborating, and the two leakage-free tasks (classification on a disjoint cohort, and DeepPSMA segmentation, absent from pre-training) show the same ranking, supporting our claim that the advantage is not a memorisation artefact. Second, our encoder uses patch size 16 against the baselines' 14 in accordance with DINOv1 and DINOv3, but per the DINO literature a larger patch should if anything weaken performance, meaning it does not explain its lead. Third, our PET pre-training set is comparatively small and MIP-based; future work includes slice-based pre-training to increase 2D diversity and remove the mismatch with the segmentation input.

Overall, the pre-training domain matters for PET: its functional signal less well captured by natural-image or CT/MRI pre-training, and an in-domain PET DINOv2 transfers better across tasks despite far less pre-training data.

\newpage
\bibliographystyle{splncs04}
\bibliography{CaPTion_2026.bib}
\end{document}